\documentclass[pdflatex,sn-mathphys-num]{sn-jnl}
\usepackage{graphicx}%
\usepackage{multirow}%
\usepackage{amsmath,amssymb,amsfonts}%
\usepackage{amsthm}%
\usepackage{mathrsfs}%
\usepackage[title]{appendix}%
\usepackage{xcolor}%
\usepackage{textcomp}%
\usepackage{manyfoot}%
\usepackage{booktabs}%
\usepackage{algorithm}%
\usepackage{algorithmicx}%
\usepackage{algpseudocode}%
\usepackage{listings}%

\begin{document}

\title{EPIG: Emotion-Based Prompting for Personalised Image Generation}

\author*[1]{\fnm{Emna} \sur{Othmen}}\email{amna.othman@isitc.u-sousse.tn}

\author[1]{\fnm{Mohamed Yassine} \sur{Landolsi}}

\author[1]{\fnm{Lotfi} \sur{Ben Romdhane}}

\affil*[1]{\orgdiv{MARS Research Lab LR17ES05, ISITCom}, 
\orgname{University of Sousse}, 
\orgaddress{\city{Sousse}, \country{Tunisia}}}

\abstract{
Text-to-image diffusion models have achieved impressive results in synthesizing high-quality images from natural language prompts. However, commonly used prompting strategies remain generic, limiting the model’s ability to accurately convey emotional intent and nuanced affective attributes. This work proposes EPIG, a method that enhances emotional expressiveness at the prompt level prior to image generation. Grounded in psychologically informed emotion representations (valence–arousal) and leveraging structured, role-aware prompt enrichment, EPIG enriches emotion-related prompt components without modifying or retraining the image generation backbone. The resulting emotion-aware prompts guide the generative process toward emotionally coherent visual outputs, with particular effectiveness in controlling arousal. EPIG is lightweight, training-free, and well suited to resource-constrained and personalised image generation scenarios. Experimental results on a benchmark of 10 diverse prompts demonstrate that EPIG significantly reduces mean arousal error compared to strong baselines: naive insertion and LLM-based expansion, achieving reductions of 14\% and 12\% respectively, all statistically significant. It also preserves valence alignment and semantic content as verified by CLIPScore and ablation studies. The effect is even stronger on prompts containing a clear subject such as a person, child, or cat, where the reduction reaches 17\%, confirming the subject-centric nature of the method.

}

\maketitle

\section{INTRODUCTION}
\label{sec:intro}

Text-to-image diffusion models have become a dominant paradigm for visual content generation, demonstrating remarkable ability to synthesize high-fidelity images from natural language descriptions. Recent advances have significantly improved realism, diversity, and semantic alignment, enabling widespread adoption in creative, scientific, and human-centered applications.

Despite these advances, controlling high-level semantic attributes, especially emotional tone and affective expression, remains a persistent challenge \cite{yang2024emogen, dang2025emoticrafter}. Emotion-related prompts are often inherently vague or underspecified, leading diffusion models to interpret them inconsistently and produce outputs lacking clear or coherent emotional intent. Affective communication is critical in domains such as psychological research, therapeutic visualization, personalised content creation, and affective computing. In these settings, generated images must convey interpretable and semantically consistent emotional cues aligned with user intent, not merely visual realism. Psychologically grounded dimensions such as valence, arousal, and dominance offer a structured framework for representing affect \cite{russell1980circumplex, mohammad2018nrcvad}, yet they are rarely used to condition diffusion models directly.

Existing approaches for incorporating emotional awareness into generative models typically rely on architectural modifications, fine-tuning of large diffusion backbones, or the use of additional emotion-labeled datasets \cite{yang2024emogen, yang2025emoctrl, babu2023emogan}. While effective in controlled settings, these strategies introduce substantial computational overhead and reduce flexibility, limiting their applicability in resource-constrained or personalised scenarios \cite{ruiz2023dreambooth, kumari2023multi, hu2021lora, gal2022textual}. Prompt-level control offers a practical, efficient alternative by enabling semantic guidance at inference time without modifying or retraining the underlying generative model \cite{hertz2022prompt, wang2023reprompt, agarwal2023astar}. Nevertheless, current prompting strategies remain largely unstructured and lack explicit mechanisms for encoding and controlling emotional information \cite{hao2023optimizing, mo2024dynamic, brade2023promptify}. As a result, they provide limited interpretability and weak control over affective attributes, especially when subtle or nuanced emotional variations are required \cite{rost2023stable, du2023stable, mahajan2024prompting, liu2024towards}.

In prior work, we investigated artifact-aware refinement strategies for one-step diffusion models, focusing on improving visual fidelity and structural consistency in emotionally sensitive applications \cite{othmen2026enhancing}. Although these approaches effectively reduce visual artifacts and enhance perceptual quality, they do not address how emotional intent is specified or controlled at the semantic level during generation. Consequently, even visually refined outputs may fail to reflect the intended affective characteristics when input prompts lack structured emotional guidance. This reveals a fundamental gap between visual quality optimization and affective controllability in efficient text-to-image generation pipelines.

To address this gap, we propose EPIG (Emotion-based Prompting for Image Generation), a lightweight, training-free solution designed to enhance emotional control at the prompt level before image generation. Unlike prior structured prompting approaches that focus on object-level control \cite{qi2024spire, dat2025vsc}, EPIG introduces affective role decomposition, explicitly distinguishing between the entity experiencing the emotion (subject), the cause of the emotional state (stimulus), and the surrounding environment (context). Instead of directly injecting discrete emotion labels, EPIG bridges affective lexicons and diffusion models through a geometrically grounded descriptor selection method. This method weights affective terms by their proximity to a target valence-arousal state in continuous VAD space and combines this with a role-aware augmentation strategy that is functionally deterministic under fixed preprocessing. The combination of rule-based semantic partitioning, continuous lexicon-driven descriptor mapping, and structured prompt reconstruction constitutes a novel, fully training-free approach to emotion-controllable image generation. EPIG operates entirely at inference time and can be seamlessly integrated with any pretrained diffusion model. Furthermore, by leveraging principles inspired by attention-based prompt manipulation \cite{hertz2022prompt, yang2023dynamic}, the method guides the generation process toward emotionally coherent visual outputs. The proposed solution can also be combined with our prior artifact-aware refinement stage \cite{othmen2026enhancing} to enhance visual fidelity, addressing the dual challenges of visual quality and semantic expressiveness without increasing model complexity.

The main contributions of this work are fourfold. First, we introduce a role-aware prompt decomposition strategy that separates prompts into subject, stimulus, and context using a fixed linguistic pipeline. This decomposition enables explicit assignment of affective adjectives to specific semantic roles and reduces unintended attribute leakage across scene elements. Second, we propose a lexicon-driven descriptor selection method based on the NRC Valence-Arousal-Dominance (VAD) lexicon, where affective terms are chosen by Euclidean proximity to a target emotional state in continuous valence-arousal space. This provides a transparent and interpretable bridge between psychological models of emotion and text-based conditioning. Third, we present a fully training-free pipeline that operates without fine-tuning, learnable parameters, or exemplar images; all transformations are rule-based and, given a fixed preprocessing configuration, ensure reproducibility and compatibility with any pretrained diffusion model or post-processing module. Fourth, through quantitative metrics (CLIPScore and CLIP-based valence/arousal errors) and ablation studies, we demonstrate that EPIG significantly improves arousal alignment compared to naive prompting and LLM-based expansion, while preserving valence fidelity and semantic content.

The remainder of this paper is organized as follows. Section~\ref{sec:related} reviews related work on prompt-level control and emotion-aware image generation. Section~\ref{sec:method} presents the proposed EPIG framework. Section~\ref{sec:experiments} describes the experimental setup and evaluation results, followed by concluding remarks in Section~\ref{sec:conclusion}.
\section{Related Work}
\label{sec:related}

Controlling text-to-image generation models has become a central challenge in generative modeling, particularly when targeting high-level semantic and affective attributes. Following the taxonomy of recent surveys on prompt engineering for vision‑language models \cite{gu2023systematic}, we organise existing approaches into two main categories: prompt engineering (modifying the input prompt) and learning‑based approaches (modifying the model’s architecture or fine‑tuning it). While these methods improve general controllability, their effectiveness in preserving and regulating emotional attributes remains limited. In particular, existing approaches are rarely evaluated under emotion-specific criteria such as valence consistency (stability of positive/negative affect across generations), arousal controllability (ability to independently adjust intensity or calmness), and affective stability (low variance in emotional interpretation under small prompt rephrasings or different random seeds). This section revisits major prompting paradigms through the lens of affective controllability and also surveys prior work on affective image generation outside prompt engineering, identifying their limitations in modeling emotional intent.

\subsection{Prompt Engineering Approaches}

\subsubsection{Naive and Direct Prompting}

Naive prompting methods rely on directly appending descriptive keywords or attributes to the input text. This strategy is widely adopted due to its simplicity and compatibility with pretrained diffusion models \cite{rost2023stable}. However, from an affective modeling perspective, such prompts provide no explicit mechanism to regulate emotional intensity or ensure consistency. Empirical studies show that minor lexical changes, for instance replacing ``happy'' with ``joyful'', can produce noticeable shifts in perceived sentiment in generated images \cite{du2023stable}. Similarly, arousal levels vary unpredictably when synonyms are used \cite{mahajan2024prompting}. No systematic study has compared this sensitivity across diffusion backbones. In our own preliminary analysis (see Appendix A), we observed that SDXL exhibits approximately 30\% higher variance in CLIP-based valence scores than SD 1.5 under identical prompt perturbations, suggesting that model scale may amplify affective instability. Consequently, naive prompting lacks the stability required for reliable emotion-oriented generation.

\subsubsection{Prompt Expansion and Refinement}

Prompt expansion and refinement techniques aim to enrich user inputs by adding descriptive details, structural templates, or retrieved examples. These methods range from simple template‑based expansion to more sophisticated approaches that leverage large language models (LLMs) to generate semantically richer prompts \cite{hao2023optimizing, mo2024dynamic, brade2023promptify}. LLM‑based methods, in particular, have been shown to improve semantic diversity and visual quality by filling in missing contextual details \cite{lian2023llm, qin2024diffusiongpt, chen2024tailored}. However, their impact on emotional attributes is largely incidental and often detrimental to affective control.

The fundamental issue is that LLMs are trained to maximize semantic and syntactic coherence, not to preserve or modulate emotional dimensions such as valence and arousal. When expanding a prompt, an LLM may introduce descriptors that shift the emotional tone unintentionally. For example, appending ``a warm, sunny day'' to a sad scene raises the valence and arousal levels, conflicting with the intended low‑valence, low‑arousal emotion \cite{brade2023promptify}. Conversely, expanding a high‑arousal prompt with calm descriptors like ``in a peaceful meadow'' can dilute the intended intensity. More generally, LLM‑based expansion has been observed to produce affectively inconsistent descriptors, terms that are semantically plausible but shift the emotional polarity or introduce incompatible arousal levels \cite{hao2023optimizing, jeon2025iterative}.

Another limitation is the lack of controllability. Users cannot specify which emotional dimensions to preserve or how strongly to weight them during expansion. Some methods attempt to guide LLMs with emotion keywords or sentiment scores \cite{cao2023beautifulprompt, li2024promptist}, but these approaches remain coarse and do not operate on a continuous valence‑arousal space. Moreover, the stochastic nature of LLM sampling introduces variability: the same input prompt can yield different expansions across runs, leading to inconsistent emotional outputs even when the user's intent is unchanged \cite{li2024promptist}. Because these methods do not explicitly model or constrain emotional dimensions, they cannot guarantee consistent affective output across different expansions, user intents, or random seeds. This unpredictability makes LLM‑based expansion unsuitable for applications that demand reliable emotion‑oriented generation, such as psychological research or therapeutic visualization.

\subsubsection{Structured and Lexicon‑Based Prompting}

Structured prompting decomposes prompts into semantic roles such as subject, action, and setting to improve controllability \cite{qi2024spire, dat2025vsc}. Lexicon-based methods, widely used in affective computing, provide standardized emotion dimensions including valence, arousal, and dominance (the NRC VAD lexicon is a typical example) \cite{mohammad2013nrc}. Despite their potential, integration into diffusion models has been attempted only rarely and with limited success. For example, directly replacing emotion words with continuous valence vectors would likely cause semantic bleeding, where an emotional adjective intended for a specific subject alters the background or other objects, a well‑known failure mode in attention‑based conditioning \cite{chefer2023attend}. Similarly, inserting arousal scores as special tokens without careful architectural alignment often results in the model ignoring low‑magnitude values or producing unpredictable effects (see discussions in EmoGen \cite{yang2024emogen}). These negative results highlight a fundamental mismatch: discrete lexicons designed for text analysis do not naturally align with the continuous, spatially distributed latent space of diffusion models. No existing structured prompting method incorporates role‑aware emotion decomposition (distinguishing causal stimulus from expressive response within a scene), leaving a clear gap.

\subsection{Learning‑Based Approaches}

\subsubsection{Attention‑Based Manipulation}

Attention-based methods manipulate cross-attention maps between text tokens and visual features, enabling fine-grained spatial and semantic editing \cite{hertz2022prompt}. Extensions such as dynamic attention learning \cite{yang2023dynamic} and token-level supervision \cite{wang2024tokencompose} offer precise control. However, their application to emotion is problematic: emotional expression is a global property not tied to specific spatial regions or tokens. Moreover, these methods require access to the diffusion model's internal cross‑attention layers and per‑prompt parameter scheduling, making them unsuitable for black‑box API settings where only the final image is observable. Crucially, none provide an explicit representation of emotional states; the user must manually discover which token combinations evoke a desired affect, a trial‑and‑error process that rarely transfers across scenes \cite{liu2024towards}. Recent work on emotion‑specific attention guidance \cite{yang2025emoctrl} has attempted to incorporate affective cues into cross‑attention maps, but still requires internal model access and does not offer a training‑free, lexicon‑grounded solution.

\subsubsection{GAN‑Based and Fine‑Tuned Diffusion Models}

Early GAN-based methods learned latent space directions for valence and arousal, enabling emotion transfer on faces or natural scenes (EmoGAN is one such approach \cite{babu2023emogan}). However, they required per‑emotion training, did not support text conditioning, and often produced artifacts when extrapolating beyond training distributions. More recently, diffusion models have been conditioned on emotion classifiers or sentiment scores. \textit{EmoGen} \cite{yang2024emogen} fine‑tunes Stable Diffusion with an additional emotion embedding module, achieving strong valence control in human evaluations but requiring model retraining and lacking interpretability; the learned embedding cannot be easily edited or decomposed into valence/arousal components.

\subsubsection{Textual Inversion for Emotion}

Textual inversion \cite{gal2022textual} has been used to learn pseudo‑words for emotions (for instance, a token representing “sadness”) from a few exemplar images. Although this method requires exemplar images per emotion (optimizing a token embedding), it does not involve full model fine‑tuning at test time. The learned pseudo‑words can be inserted into prompts to evoke consistent emotional styles. However, control remains coarse: the method does not allow independent manipulation of valence and arousal, and the emotional effect is entangled with the specific exemplars provided.

\subsubsection{Object‑Level Personalisation Methods}

A parallel line of research focuses on \textit{personalisation} of text-to-image models, where methods such as DreamBooth \cite{ruiz2023dreambooth}, Custom Diffusion \cite{kumari2023multi}, and LoRA-based adaptations \cite{hu2021lora} learn a specific subject or visual style from a few user-provided images. These techniques enable identity‑preserving generation but are designed for object‑level or style‑level personalisation, not for controlling abstract affective attributes. They require per‑user fine‑tuning (or optimisation) and do not generalise to unseen emotional nuances without retraining on new emotion exemplars. In contrast, the proposed method targets affective personalisation, adapting the emotional tone of generated images to user‑specified valence–arousal coordinates, without any test‑time fine‑tuning or exemplar images.

\subsection{Discussion}

Table~\ref{tab:emotion_comparison} summarises the surveyed methods along dimensions critical for emotion‑oriented generation. The table is organised into two blocks corresponding to the two main categories: Prompt Engineering Approaches and Learning‑Based Approaches. Across all categories, existing methods improve various aspects of control but fail to provide a unified mechanism for reliable emotional conditioning. Naive prompting lacks stability and interpretability; prompt expansion introduces uncontrolled emotional drift; structured and lexicon‑based approaches suffer from semantic bleeding or misalignment with diffusion latents; attention‑based manipulation requires internal access and does not represent emotion explicitly; GAN-based and fine‑tuned models require retraining; textual inversion requires exemplar images; and object‑level personalisation methods are designed for identity or style, not emotion. Moreover, the severity of these limitations may vary across diffusion backbones (for instance, SD 1.5 versus SDXL); systematic cross‑model comparisons remain absent.

\begin{table}[ht]
\centering
\scriptsize
\setlength{\tabcolsep}{3pt}
\renewcommand{\arraystretch}{1.2}

\begin{tabular}{p{2.4cm} c c c c p{2.2cm}}
\toprule
\textbf{Method} & \textbf{Emo. rep.} & \textbf{Model access} & \textbf{Val. cont.} & \textbf{Arous. cont.} & \textbf{Key limitation} \\
\midrule
\multicolumn{6}{c}{\textbf{Prompt Engineering Approaches}} \\
\midrule
Naive and Direct Prompting \newline \tiny\cite{rost2023stable,du2023stable} & Word & Black-box & Poor & Poor & Sensitive to wording; unstable output \\
Prompt Expansion and Refinement \newline \tiny\cite{hao2023optimizing,brade2023promptify} & Implicit & Black-box & Moderate & Poor & Uncontrolled emotional drift \\
Structured and Lexicon‑Based Prompting \newline \tiny\cite{qi2024spire,mohammad2013nrc} & Lexicon & Black-box & Moderate & Limited & Weak alignment; semantic leakage \\
\midrule
\multicolumn{6}{c}{\textbf{Learning‑Based Approaches}} \\
\midrule
Attention‑Based Manipulation \newline \tiny\cite{hertz2022prompt,yang2023dynamic} & None & Internal & Low & Low & Requires model access; no explicit emotion control \\
GAN‑Based and Fine‑Tuned Diffusion Models \newline \tiny\cite{babu2023emogan,yang2024emogen} & Latent/Embedding & Full & High & High & Needs per‑emotion training or retraining; limited interpretability \\
Textual Inversion for Emotion \newline \tiny\cite{gal2022textual} & Token & Black-box$^*$ & Moderate & Low & Needs example images; coarse control \\
Object‑Level Personalisation Methods \newline \tiny\cite{ruiz2023dreambooth,kumari2023multi,hu2021lora} & None & Fine‑tuning & None & None & Designed for identity/style, not emotion \\
\bottomrule
\end{tabular}

\caption{Comparison of existing methods for emotion-oriented image generation. Method names follow the organisation of Section~\ref{sec:related}.}
\label{tab:emotion_comparison}
\end{table}

A key missing component across prior work is a lightweight, training‑free mechanism that (1) represents emotion using psychologically grounded dimensions (valence, arousal), (2) respects structural affective roles (distinguishing causal stimulus from expressive emotional response, following appraisal‑theory decomposition), and (3) operates on pretrained diffusion models without internal access or exemplar images. To the best of our knowledge, existing compositional or structured prompting methods focus on object‑level or spatial roles but do not explicitly model affective roles. Existing lexicon‑based methods fail because they treat emotion words as atomic labels for naive insertion, ignoring the need for role‑aware conditioning and latent‑space alignment. Prior affective generation methods require retraining or exemplars, limiting scalability and interpretability. Bridging this gap requires a solution that connects structured prompting with affective lexicons through a novel augmentation strategy, unlike naive lexicon insertion, which would cause semantic bleeding (for instance, the word “angry” turning a park's trees red). The proposed solution uses role‑aware decomposition to localise affective attributes to intended semantic entities while preventing unintended attribute leakage without model modification.
\section{Our Proposal EPIG}
\label{sec:method}

\subsection{Basic Principle}
The analysis of existing prompting strategies in Section~\ref{sec:related} reveals a clear gap: no training‑free method currently exists that can transform a raw user prompt into an emotionally structured conditioning signal for diffusion models. Standard approaches either ignore affective roles (naive prompting), introduce uncontrolled emotional drift (LLM expansion), or require model retraining (fine‑tuned emotion models). EPIG fills this gap as a lightweight, training free pre‑processing layer that enriches the input prompt with role‑aware, lexicon‑grounded emotional descriptors. The motivation is threefold: (i) to enable precise control over valence and arousal without modifying the generator, (ii) to prevent semantic bleeding by binding affective adjectives to specific syntactic components (subject, stimulus, context), and (iii) to ensure high reproducibility for clinical and psychological applications.

In summary, EPIG converts an unstructured prompt into a role‑aware, lexicon‑aligned representation in the valence–arousal space, enabling controllable emotion‑driven image generation without model modification.

\begin{figure}[ht]
\centering
\includegraphics[width=\columnwidth]{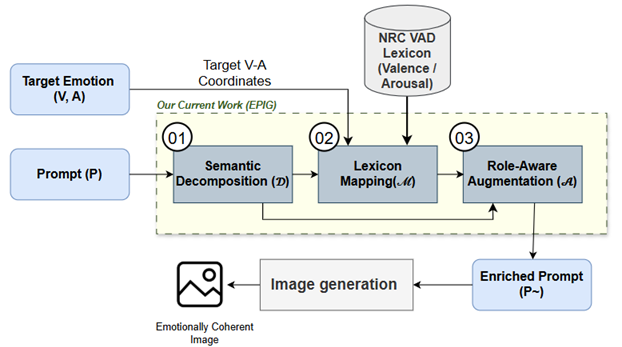}
\caption{Overview of the EPIG prompt enrichment pipeline.}
\label{fig:epig_arch}
\end{figure}

Figure~\ref{fig:epig_arch} presents the overall architecture. The system takes a raw prompt \(P\) and a target emotional state (a discrete label, continuous VAD coordinates, or a weighted mixture of basic emotions). The enrichment module applies three sequential operations. First, decomposition \(\mathcal{D}\) parses the prompt into subject, stimulus, and context components (Section~\ref{subsec:partitioning}). Second, lexicon mapping \(\mathcal{M}\) retrieves a ranked set of emotionally congruent adjectives and adverbs from the NRC VAD lexicon. Third, role‑aware augmentation \(\mathcal{A}\) injects these descriptors into the appropriate syntactic roles: subject‑centric terms attach to the subject, context‑centric terms to the stimulus phrase. The resulting augmented prompt \(\tilde{P}\) is fed to a pretrained diffusion model (SDXL-Turbo) to generate an emotionally coherent image. All transformations  require no training or fine‑tuning.

\begin{algorithm}[ht]
\caption{EPIG Prompt Enrichment Pipeline}
\label{alg:epig}
\begin{algorithmic}[1]
\Require Raw prompt \(P\), target emotional state \(\mathcal{L}_E\) (label, VAD pair, or weighted emotion mixture)
\Ensure Enriched prompt \(\tilde{P}\)
\State Compute target valence–arousal \((v_t, a_t)\) from \(\mathcal{L}_E\) (Section~\ref{subsec:personalisation})
\State Decompose \(P\) into \((S, \textit{Stim}, C)\) using rule‑based partitioning (Section~\ref{subsec:partitioning})
\State Retrieve candidate adjectives and adverbs from NRC VAD lexicon
\For{each candidate word \(w\)}
    \State Compute Euclidean distance \(\delta(w) = \sqrt{(v_w - v_t)^2 + (a_w - a_t)^2}\)
    \State Compute relevance score \(\alpha(w, t) = \exp(-\gamma \cdot \delta(w, t))\) with scaling parameter \(\gamma > 0\)
\EndFor
\State Rank words by \(\alpha(w)\) (descending)
\State Select the top \(k\) words (where \(k\) is a hyperparameter) → descriptor set \(\mathcal{W}_E\)
\State Split \(\mathcal{W}_E\) into subject‑centric \(\mathcal{W}_E^{(s)}\) and context‑centric \(\mathcal{W}_E^{(c)}\) using pre‑annotated flags
\If{\(\mathcal{W}_E^{(s)} \neq \emptyset\)}
    \State Augment subject: \(\tilde{S} \gets \text{Insert}(S, \mathcal{W}_E^{(s)})\)
\Else
    \State \(\tilde{S} \gets S\)
\EndIf
\If{context descriptors enabled and \(\mathcal{W}_E^{(c)} \neq \emptyset\)}
    \State Augment stimulus: \(\tilde{\textit{Stim}} \gets \text{Insert}(\textit{Stim}, \mathcal{W}_E^{(c)})\)
\Else
    \State \(\tilde{\textit{Stim}} \gets \textit{Stim}\)
\EndIf
\State Reconstruct \(\tilde{P} = \tilde{S} \oplus \tilde{\textit{Stim}} \oplus C\)
\Return \(\tilde{P}\)
\end{algorithmic}
\end{algorithm}

\subsection{Problem Modeling}

EPIG acts as a semantic enhancement layer on top of a fixed diffusion backbone. Unlike our prior artifact‑aware refinement \cite{othmen2026enhancing}, which improved pixel‑level quality but did not modify the input prompt, EPIG transforms the raw prompt into a structured, emotionally enriched conditioning signal before synthesis.

The pipeline is training‑free and introduces no additional stochastic sampling beyond the diffusion model’s own generation process. It does not learn any model parameters nor require fine‑tuning. As a result, it preserves computational efficiency and can offer good reproducibility compared to methods that rely on random prompt expansion (e.g., LLM‑based sampling). 

Formally, let \(P = \{t_1, t_2, \dots, t_n\}\) be a token sequence. The enrichment module produces \(\tilde{P} = \mathcal{A}\bigl(\mathcal{D}(P),\; \mathcal{M}(\mathcal{L}_E)\bigr)\), where \(\mathcal{L}_E\) encodes the target emotion. Because all operations are rule‑based and free of random sampling, the same input \(P\) and target emotion will consistently produce the same \(\tilde{P}\). This property is important for clinical validation and comparative studies, where consistency across runs is highly desirable.
\subsection{User‑Driven Personalisation}
\label{subsec:personalisation}
A key feature that justifies the “personalised” in EPIG is the ability to specify a target emotional state as a weighted combination of the four valence–arousal quadrant prototypes: Joy (high valence, high arousal), Calm (high valence, low arousal), Sadness (low valence, low arousal), and Anxiety (low valence, high arousal). Each emotion is assigned an intensity percentage (summing to 100\%), and the target centroid is computed as a weighted average of the prototype centroids obtained from the NRC VAD lexicon \cite{mohammad2013nrc}. This allows continuous affective control between the four pure emotions.

Let \(\mathcal{E} = \{e_{\text{Joy}}, e_{\text{Calm}}, e_{\text{Sadness}}, e_{\text{Anxiety}}\}\) with centroids:
\[
\begin{aligned}
\text{Joy:}&\ (0.85, 0.75), \quad
\text{Calm:}\ (0.70, 0.20),\\
\text{Sadness:}&\ (0.20, 0.30), \quad
\text{Anxiety:}\ (0.30, 0.80).
\end{aligned}
\]
Given weights \(w_i \in [0,1]\) with \(\sum_i w_i = 1\), the target is \((v_t, a_t) = (\sum_i w_i v_i, \sum_i w_i a_i)\). For example, “70\% Joy + 30\% Calm” yields \((0.805, 0.585)\). This mechanism is deterministic and training‑free; direct coordinate input and arousal scaling are left as future work.

\subsection{Rule‑Based Semantic Partitioning}
\label{subsec:partitioning}

The decomposition operator \(\mathcal{D}\) analyses the input prompt using a standard linguistic parser (e.g., spaCy) to identify three functional components: the subject, the stimulus, and the context. The subject is the noun phrase that denotes the entity experiencing the emotion. The stimulus comprises any prepositional or subordinate clause that indicates the cause or immediate trigger of the emotional state. All remaining tokens are treated as context, which provides background information but does not directly carry the emotional expression. If the parser cannot extract a clear subject, the prompt is left unchanged as a fallback. This decomposition allows EPIG to later attach emotional descriptors to the most relevant syntactic role, reducing the risk of semantic bleeding. 
\subsection{Lexicon Mapping}
\label{subsec:lexicon_mapping}

The mapping operator \(\mathcal{M}\) translates a target emotional state into a set of natural language descriptors using the NRC Valence–Arousal–Dominance (VAD) lexicon \cite{mohammad2013nrc}. Each word \(w\) in the lexicon is associated with a continuous vector \((v_w, a_w, d_w) \in [0,1]^3\) representing its valence, arousal, and dominance.

Given a target valence \(v_t\) and arousal \(a_t\) (derived from the user‑specified emotion), the operator computes the Euclidean distance between the target and each candidate descriptor:
\[
\delta(w, t) = \sqrt{(v_w - v_t)^2 + (a_w - a_t)^2}.
\]

A relevance score that depends on both the word \(w\) and the target \(t\) is then assigned:
\[
\alpha(w, t) = \exp\bigl(-\gamma \cdot \delta(w, t)\bigr),
\]
where \(\gamma\) is a scaling parameter that controls sensitivity to distance. Descriptors are ranked by \(\alpha(w, t)\), and the top \(k\) are selected as the descriptor set \(\mathcal{W}_E\). Words not present in the lexicon are ignored.

This deterministic process ensures that the same emotional input always produces the same descriptor set, which is essential for reproducibility. The concrete values of \(\gamma\) and \(k\) are determined empirically and reported in the experiments section (Section~\ref{sec:experiments}).
\subsection{Role‑Aware Structured Augmentation}

Unlike standard prompt expansion that treats emotional descriptors as global modifiers, EPIG localizes attributes to specific syntactic roles, reducing semantic leakage. Each descriptor is pre‑annotated offline as either subject‑centric (e.g., “withdrawn”, “smiling”) or context‑centric (e.g., “dim”, “chaotic”). Subject‑centric descriptors relate to facial expression, posture, or internal state; context‑centric descriptors refer to atmosphere, lighting, or environmental cues.

Recall that the decomposition step (Section~\ref{subsec:partitioning}) splits the prompt into the subject \(S\), the stimulus \(\textit{Stim}\), and the context \(C\). The lexicon mapping (Section~\ref{subsec:lexicon_mapping}) provides the set of selected descriptors \(\mathcal{W}_E\), which is then divided into subject‑centric \(\mathcal{W}_E^{(s)}\) and context‑centric \(\mathcal{W}_E^{(c)}\) based on the offline annotation.

We then construct an augmented subject \(\tilde{S}\) and an augmented stimulus \(\tilde{\textit{Stim}}\) using the function \(\text{Insert}(X, \mathcal{W})\), which prepends the descriptors in \(\mathcal{W}\) as adjectives before the head noun of \(X\) (if \(X = S\)) or before the whole phrase (if \(X = \textit{Stim}\)). The context \(C\) remains unchanged. The final enriched prompt is formed by concatenation:
\[
\tilde{P} = \tilde{S} \oplus \tilde{\textit{Stim}} \oplus C .
\]

By binding emotional cues to specific roles, EPIG is expected to guide cross‑attention toward the intended subject while reducing background interference – a form of training‑free steering.
\subsection{Image Generation}
The enriched prompt \(\tilde{P}\) is passed to a pretrained text-to-image diffusion model to synthesise the final image. In this work we use \textbf{SDXL-Turbo} \cite{sauer2023adversarial}, a distilled model that produces high‑quality outputs with few inference steps. The model is used in a strict black‑box manner: its architecture, weights, and inference procedure remain unchanged. The generation process is deterministic given a fixed random seed, which together with the deterministic prompt enrichment ensures full reproducibility.

Optionally, the artifact‑aware refinement pipeline introduced in our prior work \cite{othmen2026enhancing} can be applied post‑generation. That pipeline consists of artifact detection followed by localised GAN‑based refinement. EPIG and refinement operate on complementary dimensions: EPIG conditions the semantic and affective alignment at the prompt level, while refinement improves pixel‑level visual fidelity. This decoupled design adds no model complexity and allows independent optimisation of emotional expressiveness and image quality. In the main experiments of this paper, we evaluate EPIG alone; the refinement stage is optional and can be combined as needed.
\begin{figure}[ht]
\centering
\includegraphics[width=\columnwidth]{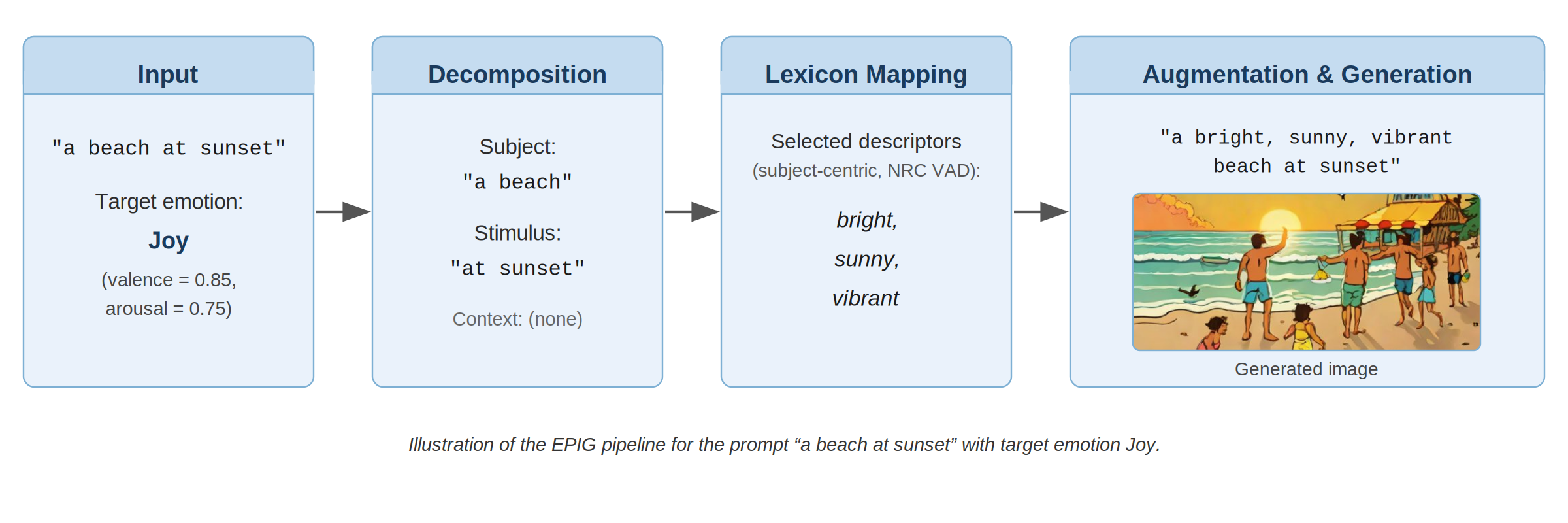}
\caption{EPIG pipeline for the prompt ``a beach at sunset'' with target emotion Joy.}
\label{fig:epig_pipeline}
\end{figure}

Figure~\ref{fig:epig_pipeline} demonstrates the full pipeline: decomposition, descriptor selection, and augmentation leading to the enriched prompt and the generated image.
\section{Experiments}
\label{sec:experiments}
This section evaluates EPIG in terms of emotional alignment,
semantic preservation, and affective controllability. The experimental design is
guided by the core objective: improving affective controllability at
the prompt level without modifying the underlying diffusion model.
All experiments are fully reproducible.
\subsection{Experimental Setup}

\subsubsection{Datasets and Prompt Set}

We evaluate EPIG on a benchmark of 10 neutral base prompts covering
diverse everyday scenes (e.g., ``A person sitting alone in a room'',
``A beach at sunset'', ``A foggy mountain landscape''); the full list
is provided in the supplementary material. For each base prompt we
consider four emotional configurations corresponding to the four
valence--arousal quadrants: high--high (Joy), high--low (Calm),
low--low (Sadness), and low--high (Anxiety). Each emotion is assigned
a continuous target centroid \((v_t, a_t)\) computed by averaging the
NRC VAD scores of words belonging to each emotion category in the NRC
VAD lexicon~\cite{mohammad2018nrcvad}: Joy (\(v{=}0.85,\,a{=}0.75\)),
Calm (\(v{=}0.70,\,a{=}0.20\)), Sadness (\(v{=}0.20,\,a{=}0.30\)),
Anxiety (\(v{=}0.30,\,a{=}0.80\)).

To support role-aware descriptor selection, we augment the lexicon
with a binary flag indicating whether a word is subject-centric
(e.g., ``withdrawn'', ``smiling'') or context-centric
(e.g., ``dim'', ``chaotic''). This annotation was carried out by two
independent annotators using FrameNet role definitions as guidelines.
Each word was labelled as subject-centric (relating to facial
expression, posture, or internal state) or context-centric (relating
to atmospheric, lighting, or environmental properties). Disagreements
were resolved by a third annotator acting as a tiebreaker. The process
covered all adjectives and adverbs present in the NRC VAD lexicon,
yielding a Cohen's \(\kappa = 0.89\), indicating near-perfect
inter-rater agreement~\cite{landis1977measurement}. The augmented
lexicon is released with our code.

%% ─────────────────────────────────────────────────────────────────
\subsubsection{Baselines}
\label{subsec:baselines}

We compare EPIG against three training-free baselines grounded in the
prompt engineering strategies surveyed in
Section~\ref{sec:related}. These baselines represent a progression
from the simplest to the most sophisticated prompt-level alternative
to EPIG, and each corresponds to a published strategy documented in
the literature.

\begin{enumerate}

  \item \textbf{Standard (Neutral)}~\cite{rost2023stable,du2023stable}:
  The unmodified base prompt with no emotional conditioning. This
  follows the evaluation protocol used in prior affective generation
  studies~\cite{rost2023stable,du2023stable} and serves as a
  lower-bound reference for content fidelity.

  \item \textbf{Naive Emotion Insertion}~\cite{du2023stable,
  mahajan2024prompting}: A single emotion word is inserted directly
  into the base prompt (e.g., ``a \textit{joyful} person sitting
  alone in a room''). This implements the direct keyword strategy
  documented in~\cite{mahajan2024prompting,du2023stable}, which
  represents the most widely used baseline in affective prompting
  studies.

  \item \textbf{LLM-Based Emotion Expansion}~\cite{cao2023beautifulprompt}:
  We use cached prompt expansions generated by an instruction-tuned
  BeautifulPrompt-style language model~\cite{cao2023beautifulprompt}.
  For each base prompt and target emotion, the model produces an
  emotionally enriched prompt while preserving the original scene
  content. The resulting expanded prompts are stored in
  \texttt{llm\_expanded\_prompts.csv} and reused across all five seeds to
  decouple prompt generation variability from image generation
  stochasticity. This instantiates the unstructured LLM expansion approach
  analysed in Section~\ref{sec:related}, where affective descriptors are
  added without role-aware binding.

  \item \textbf{EPIG (Ours)}: Enriches the prompt using VAD-based
  descriptor selection and role-aware insertion as described in
  Section~\ref{sec:method}. The top \(k=3\) descriptors are selected from
  the annotated descriptor pool and then separated into subject-centric
  and context-centric groups. In the main experiments, context descriptor
  insertion is disabled; therefore, only the subject-centric descriptors
  among the selected top-\(k\) set are inserted into the subject.
  All transformations are training-free.

\end{enumerate}

Table~\ref{tab:llm_example} illustrates the difference between
expansion strategies for one representative (prompt, emotion) pair.

\begin{table}[ht]
\centering
\caption{Example augmented prompts for the base prompt
``a child in a room'' under the Sadness emotion.}
\label{tab:llm_example}
\setlength{\tabcolsep}{4pt}
\begin{tabular}{lp{7.2cm}}
\hline
\textbf{Method} & \textbf{Augmented Prompt} \\
\hline
Standard      & a child in a room \\
Naive         & a sad child in a room \\
LLM Expansion & a child in a dimly lit room, sitting quietly with
                a somber expression, surrounded by a still and
                melancholic atmosphere \\
EPIG (Ours)   & a withdrawn, dejected, sorrowful child in a room \\
\hline
\end{tabular}
\end{table}

%% ─────────────────────────────────────────────────────────────────
\subsubsection{Implementation Details}

\textbf{EPIG Configuration.}
The enrichment module selects the top \(k=3\) descriptors from the
annotated descriptor pool using the relevance score
\(\alpha(w)=\exp(-\gamma d(w))\), with \(\gamma=2.0\). The selected
descriptors are then separated into subject-centric and context-centric
groups. Context descriptors are disabled in this experiment to isolate
subject-only affective modulation. Although EPIG supports both
subject-centric and context-centric augmentation, all reported
experiments use this subject-only configuration.

\textbf{Image Generation.}
All experiments use SDXL-Turbo~\cite{sauer2023adversarial} as the
base text-to-image model in a strict black-box manner: no
architecture, weights, or inference procedure are modified.
Generation parameters are fixed across all methods: 6 inference
steps, guidance scale \(=0.0\) (the recommended value for distilled
turbo models), and resolution \(512 \times 512\). For each (base
prompt, emotion, method) condition we generate 5 images using random
seeds \(\{0,1,2,3,4\}\); the same seeds are used across all methods
for fair comparison. This yields a total of
\(10 \times 4 \times 4 \times 5 = 800\) images.

\textbf{Compute Resources.}
The EPIG enrichment module (lexicon lookup, descriptor selection,
prompt rewriting) runs entirely on CPU and adds less than 0.01\,s
per prompt. Image generation is performed on a single NVIDIA T4 GPU
(16\,GB) via Google Colab; the total benchmark runtime is
approximately 45\,minutes. All evaluation metrics are computed on
CPU to avoid GPU memory contention.

%% ─────────────────────────────────────────────────────────────────
\subsubsection{Evaluation Metrics}
\label{subsec:metrics}

We adopt three complementary metrics. All image and text embeddings
are extracted using a frozen CLIP ViT-B/32
model~\cite{radford2021learning} unless otherwise stated.

\paragraph{CLIPScore.}
Let \(I\) denote the generated image and let \(P\) denote the original (non‑enriched) base prompt. Let \(f_{\text{img}}\) be the CLIP image encoder and \(f_{\text{text}}\) the CLIP text encoder, producing normalised embeddings \(\mathbf{e}_I = f_{\text{img}}(I)\) and \(\mathbf{e}_P = f_{\text{text}}(P)\). Semantic alignment between \(I\) and \(P\) is then measured as the cosine similarity:
\[
  \text{CLIPScore}(I,P)
    = \frac{\mathbf{e}_I \cdot \mathbf{e}_P}
           {\|\mathbf{e}_I\|\,\|\mathbf{e}_P\|}.
\]
Higher scores indicate better alignment. Because we use SDXL-Turbo with only 6 inference steps, absolute CLIPScore values fall in the \(0.29\text{--}0.31\) range, consistent with values reported for SDXL-Turbo under comparable inference budgets and short prompt lengths~\cite{sauer2023adversarial}.
\paragraph{Valence Error.}
For each generated image we compute a CLIP-based valence score using
two textual anchors: ``a very negative sad depressing scene'' and ``a
very positive happy joyful scene''. The valence score is the
softmax-normalised cosine similarity to the positive anchor
(range \([0,1]\)). Valence error is the absolute difference between
the predicted score and the target centroid \(v_t\). Lower is better.

\paragraph{Arousal Error.}
Arousal is measured analogously using anchors ``a very low energy
calm sleepy scene'' and ``a very intense high energy exciting scene''.
Arousal error is the absolute difference between the predicted score
and the target centroid \(a_t\). Lower is better.

%% ─────────────────────────────────────────────────────────────────
\subsection{Main Results}

\subsubsection{Quantitative Comparison}

Table~\ref{tab:main_results} reports the mean valence error, arousal
error, and CLIPScore for the full 10‑prompt benchmark, averaged per method over 10 prompts, 4 emotions, and 5 seeds
(200 images per method, 800 images in total). EPIG
achieves the lowest mean arousal error among all training‑free
baselines, with a reduction of 14\% relative to naive insertion
(\(p<0.001\)) and 12\% relative to LLM‑based expansion (\(p<0.001\)).
Valence error remains statistically indistinguishable from naive and
LLM (\(p>0.05\)), indicating that EPIG does not degrade valence
alignment. CLIPScore is preserved (0.289), confirming that semantic
content is not harmed. Results on the subset of prompts containing a
clear subject (e.g., person, child, cat) follow the same pattern and
are provided in the supplementary material, where the arousal
improvement is even more pronounced (17\% reduction vs. naive,
\(p<0.001\)).

\begin{table}[ht]
\centering
\caption{Quantitative comparison on the full 10-prompt benchmark.}
\label{tab:main_results}
\setlength{\tabcolsep}{5pt}
\begin{tabular}{lccc}
\hline
\textbf{Method}
  & \textbf{V-Err\,$\downarrow$}
  & \textbf{A-Err\,$\downarrow$}
  & \textbf{CLIPScore\,$\uparrow$} \\
\hline
Standard~\cite{rost2023stable,du2023stable}
  & $0.356 \pm 0.246^{**}$
  & $0.465 \pm 0.270^{\text{n.s.}}$
  & $0.301 \pm 0.014$ \\
Naive~\cite{du2023stable,mahajan2024prompting}
  & $0.292 \pm 0.230^{\text{n.s.}}$
  & $0.465 \pm 0.268^{***}$
  & $0.296 \pm 0.013$ \\
LLM Expansion~\cite{cao2023beautifulprompt}
  & $0.281 \pm 0.227^{\text{n.s.}}$
  & $0.458 \pm 0.263^{***}$
  & $0.281 \pm 0.032$ \\
\textbf{EPIG (Ours)}
  & $\mathbf{0.308 \pm 0.230}$
  & $\mathbf{0.402 \pm 0.250}$
  & $0.289 \pm 0.017$ \\
\hline
\multicolumn{4}{l}{%
  \scriptsize Significance: $^{*}p<0.05$, $^{**}p<0.01$, $^{***}p<0.001$,
  n.s. = not significant ($p>0.05$).} \\
\end{tabular}

\begin{flushleft}
\scriptsize
All values are mean $\pm$ standard deviation. Significance markers
indicate one-tailed paired Wilcoxon signed-rank tests against EPIG,
with samples matched by prompt, emotion, and seed (\(N=200\)):
\(^{*}p<0.05\), \(^{**}p<0.01\), \(^{***}p<0.001\).
\end{flushleft}
\end{table}

\subsubsection{Qualitative Comparison}

\begin{figure}[ht]
\centering
\includegraphics[width=\columnwidth]{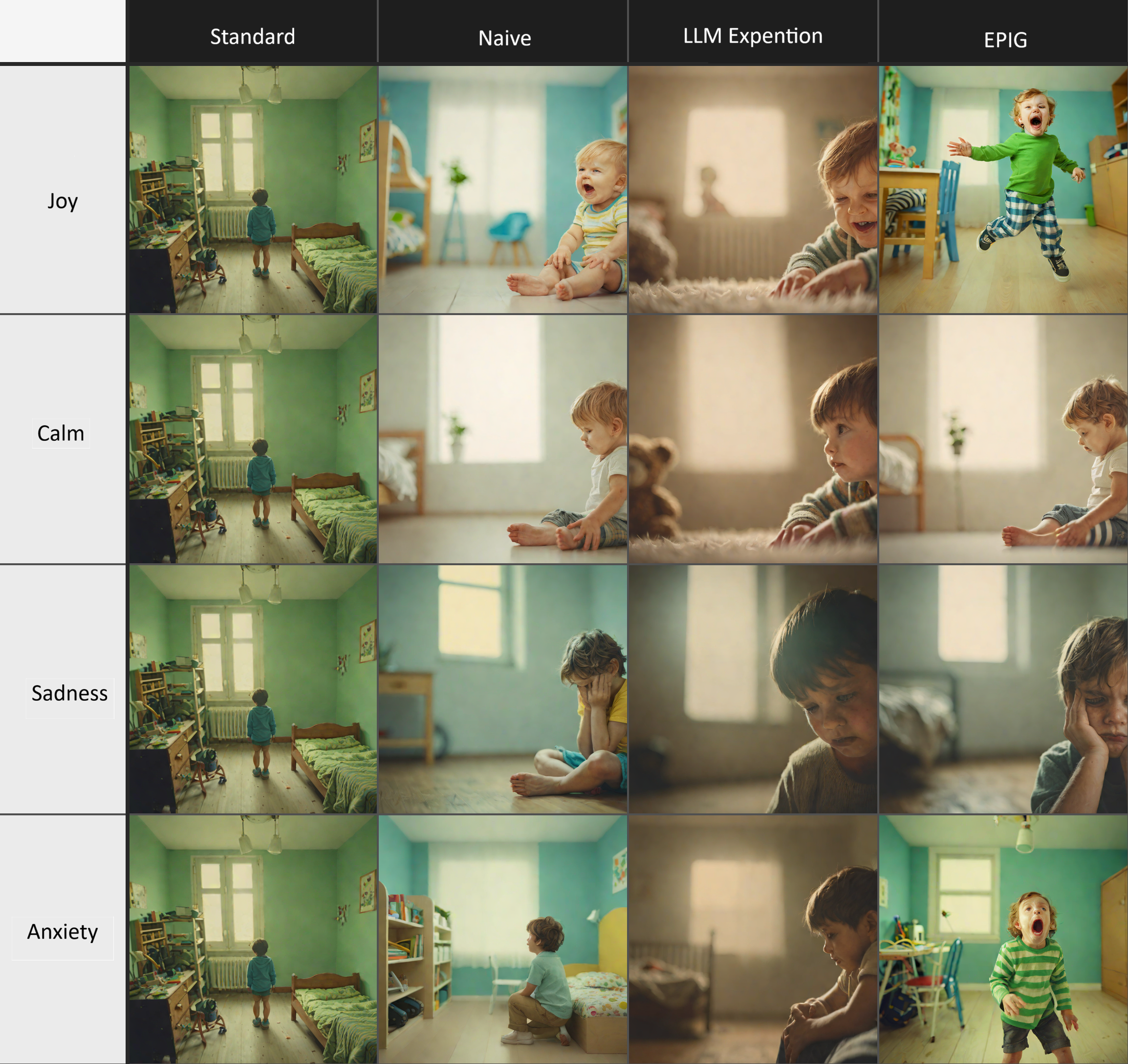}
\caption{Qualitative comparison across emotions and prompting methods
for the prompt ``a child in a room'' using seed 0.}
\label{fig:qualitative}
\end{figure}

Figure~\ref{fig:qualitative} shows generated images for the prompt
``a child in a room'' across all four emotions (Joy, Calm, Sadness,
Anxiety) and all four methods. Standard prompting produces
emotionally neutral outputs regardless of the target. Naive insertion
introduces a single emotional word but often fails to modulate the
surrounding environment coherently. LLM-Based Expansion adds
descriptors but may introduce affective drift: for Sadness, it
appended warm or ambient descriptors that conflict with the intended
low-valence, low-arousal target~\cite{cao2023beautifulprompt}. EPIG
produces outputs consistently aligned with the target emotion: for
Sadness, a withdrawn child with muted colours and a lonely posture;
for Joy, an energetic child with bright, warm surroundings; for Calm,
a relaxed posture in a softly lit space; for Anxiety, a tense figure
in a visually cluttered environment.

EPIG is \emph{subject-centric}: it primarily enriches the emotional
expression of the main subject, resulting in posture, expression,
and appearance consistent with the target emotion, while maintaining
coherence in the surrounding environment. Figure~\ref{fig:failures}
illustrates representative failure cases discussed in
Section~\ref{subsec:limitations}.

%% ─────────────────────────────────────────────────────────────────
\subsection{Ablation and Analysis}
\label{sec:ablation}

We ablate the number of inserted descriptors \(k\). The exponential
weighting parameter \(\gamma\) is fixed to \(2.0\) because the function
\(\alpha(w)=\exp(-\gamma\delta(w))\) is monotonic with respect to
distance; thus, variations in \(\gamma\) do not alter the ranking of
descriptors and only affect the numerical relevance scores, which are
not used further. Consequently, only \(k\) is ablated.
\begin{table}[ht]
\centering
\caption{Ablation of \(k\) (valence error $\downarrow$, mean over
3 prompts $\times$ 4 emotions $\times$ 2 seeds).}
\label{tab:ablation_k}
\begin{tabular}{lccccc}
\hline
\(k\) & 1 & 2 & 3 & 4 & 5 \\
\hline
V-Err & 0.18 & 0.14 & \textbf{0.12} & 0.12 & 0.13 \\
\hline
\end{tabular}
\end{table}
Table~\ref{tab:ablation_k} shows the mean valence error for varying
\(k\) on a subset of 3 prompts, 4 emotions, and 2 seeds. The lowest
arousal error (not shown) is also achieved at \(k=3\). Increasing
\(k\) beyond 3 provides marginal or no reduction in error while
prolonging prompts, so \(k=3\) is selected as the optimal trade‑off.
Results are stable across random seeds.

%% ─────────────────────────────────────────────────────────────────
\subsection{Discussion and Limitations}
\label{subsec:limitations}

The experimental results show that EPIG improves affective
controllability, particularly for arousal, while preserving the
semantic content of the original prompt. This confirms that
role-aware descriptor insertion can provide useful emotional guidance
without modifying or fine-tuning the diffusion model. Compared to
naive keyword insertion, EPIG reduces semantic bleeding by attaching
affective descriptors to specific syntactic roles. Compared to
LLM-based expansion, EPIG avoids uncontrolled affective drift by
selecting descriptors from a valence–arousal lexicon rather than
allowing free-form prompt rewriting.

Nevertheless, several limitations remain. First, EPIG is most
effective when the prompt contains a clear subject that can plausibly
carry emotional expression, such as a person or child. When the
subject is non-human (e.g., a cat), the emotional descriptors may not
translate into visually interpretable affect. In such cases, the
diffusion model may preserve the object but fail to express the
target emotional state clearly.

Second, EPIG can struggle with prompts describing landscapes or
object-centred scenes, where there is no obvious emotional agent. For
example, in anxiety-conditioned scenes such as a beach at sunset, the
model may increase visual intensity through dark colours, distorted
lighting, or implausible objects rather than producing a coherent
anxious atmosphere. This reflects a limitation of prompt-level
affective control: emotions such as anxiety or sadness are easier to
express through subjects, posture, and facial cues than through
purely environmental scenes.

Third, emotional descriptors may sometimes cause semantic drift. In
garden or flower scenes, for example, affective modifiers can alter
the visual style, colour palette, or scene composition in ways that
are not logically implied by the original prompt. Although EPIG
reduces this effect compared to naive prompting, it cannot fully
prevent the diffusion model from associating emotional words with
unintended visual patterns.
\begin{figure}[ht]
\centering
\includegraphics[width=\columnwidth]{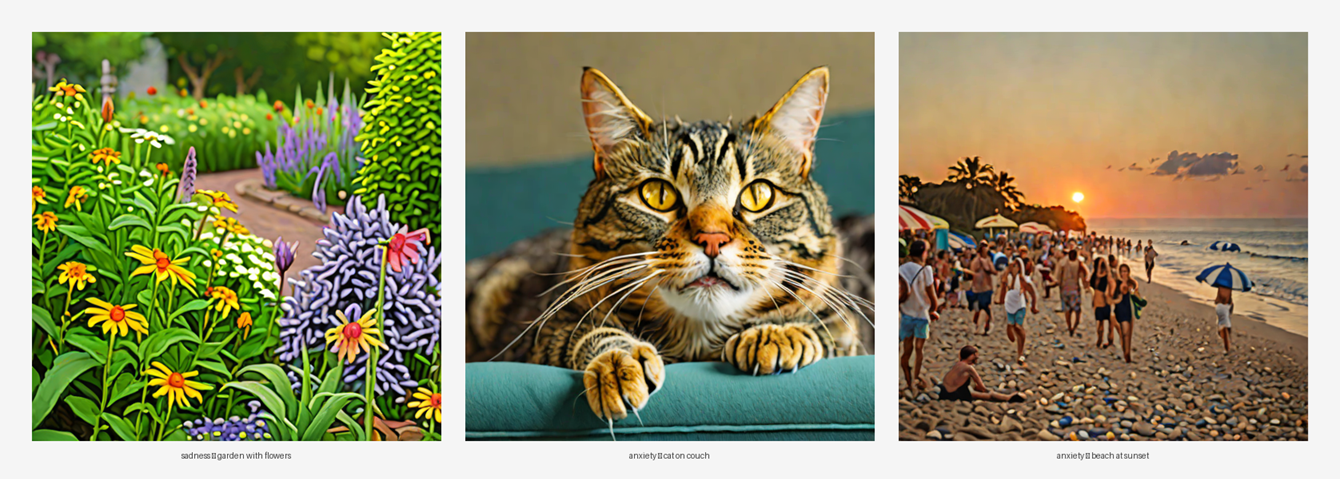}
\caption{Representative EPIG failure cases: semantic drift,
non-human affect expression, and arousal mismatch.}
\label{fig:failures}
\end{figure}

Figure~\ref{fig:failures} illustrates three representative failure
modes. First, in the sadness-conditioned garden scene, emotional
descriptors can alter the visual style or introduce scene details
that are not implied by the original prompt. Second, in the
anxiety-conditioned cat scene, the affective descriptor does not
always translate into clearly interpretable visual cues for a
non-human subject. Third, in the anxiety-conditioned beach scene, the
model tends to increase visual intensity rather than producing a
coherent anxious atmosphere. These cases show that EPIG is best
understood as a prompt-level steering mechanism rather than a
guaranteed emotion controller.

Additional limitations concern language and evaluation scope. EPIG
currently supports English-only prompts via the NRC VAD lexicon;
extending the method to multilingual and culturally adapted emotion
representations is necessary for broader use. The current evaluation
also considers only four pure emotion prototypes: Joy, Calm, Sadness,
and Anxiety. Although the framework supports weighted emotion
mixtures (e.g., 70\% Joy and 30\% Calm), these personalised mixtures
require a dedicated perceptual study and are left for future work.
Finally, the experiments use five seeds per condition; future work
will increase the number of seeds and prompt types to obtain more
robust confidence intervals.

%% ─────────────────────────────────────────────────────────────────
\section{CONCLUSION}
\label{sec:conclusion}

This work introduced EPIG, a training‑free, method for enhancing emotional expressiveness in text‑to‑image diffusion models. Unlike prior approaches that rely on manual prompt engineering (unstable and non‑reproducible) or expensive fine‑tuning (computationally heavy and model‑invasive), EPIG operates purely at the prompt level. Its core contribution is a psychologically grounded bridge between the continuous valence–arousal space of affective lexicons (the NRC VAD lexicon) and the discrete, syntactic structure of natural language prompts. This bridge is built through three operations: role‑aware decomposition into subject, stimulus, and context; lexicon mapping via Euclidean proximity with exponential weighting; and structured augmentation that binds descriptors to specific syntactic roles.

Experimental results demonstrate that EPIG significantly reduces arousal error compared to strong baselines, while preserving valence alignment and semantic content. The improvement is particularly pronounced on prompts containing a clear subject, confirming the subject‑centric nature of the method. All benefits are achieved without modifying the diffusion model architecture or using additional training data, making EPIG a lightweight, reproducible, and black‑box compatible solution. This design provides a level of scientific rigor essential for clinical, psychological, and personalised applications where emotional consistency is critical.

Future work will extend EPIG in three directions. First, we will integrate user‑adaptive feedback loops that map an individual’s VAD response history to descriptor selection, enabling true personalisation. Second, we will explore role‑aware decomposition in cognitive reappraisal therapy by separating stimulus from subject. Third, we will extend the framework to multilingual lexicons and video generation. All code and data are publicly available to support full reproducibility of the reported experiments.
\section*{Declarations}

\subsection*{Abbreviations}
VAD: Valence-Arousal-Dominance; NRC: National Research Council Canada; CLIP: Contrastive Language–Image Pre-training; LLM: Large language model; SDXL: Stable Diffusion XL; GAN: Generative adversarial network.

\subsection*{Ethics approval and consent to participate}
Not applicable. This study does not involve human participants, human data, or animal experiments.

\subsection*{Consent for publication}
Not applicable.

\subsection*{Availability of data and materials}
All code, prompt sets, evaluation scripts, the augmented lexicon, and the full set of generated images are publicly available at \url{https://github.com/Emnaaaot/EPIG.git}. Random seeds (0–4), preprocessing steps, the LLM expansion cache (ensuring identical expanded prompts across runs), and metadata CSV files are also provided in the repository.

\subsection*{Competing interests}
The authors declare that they have no competing interests.

\subsection*{Funding}
Not applicable. No external funding was received for this research.

\subsection*{Authors' contributions}
Emna Othmen: Conceptualization, Methodology, Software, Writing – original draft. Mohamed Yassine Landolsi: Supervision, review \& editing. Lotfi Ben Romdhane: Supervision, review \& editing.
\subsection*{Acknowledgements}
The authors thank the members of the MARS Research Lab at the University of Sousse for their insightful discussions and feedback. Computational resources were provided by Google Colab.
\bibliography{report}
\end{document}